RESEARCH

# Sequence-SOD: Bio-inspired Sequence-aware Spiking Object Detection for Event Cameras

**Katharina Bendig[1] · René Schuster[1,2] · Didier Stricker[1,2]**

**Abstract**

Event cameras follow a retina-inspired sensing principle, reporting local intensity changes asynchronously with high temporal resolution and a wide dynamic range. Spiking Neural Networks (SNNs) complement these sparse event streams through brain-inspired dynamics, using sparse spikes and leaky membrane potentials to integrate information over time. However, many SNN object detectors process isolated event intervals with a single label and reset the network state after each prediction, thereby underusing temporal information in continuous event streams. We introduce Sequence-SOD, a sequence-aware SNN object detector that processes extended event sequences containing labels at multiple time points. Events are accumulated into short intervals, discretized into temporal steps, and fed sequentially to an SSD-style Spiking DenseNet while preserving membrane potentials across intervals within a sequence, so that detection is driven by an evolving neural state instead of independently reset input windows. On the Gen1 Automotive Detection Dataset, sequence-aware training improves mAP from 23.38 for single-interval training to 25.30 without augmentation and to 26.88 with event-data augmentation. The model achieves a theoretical prediction frequency of 40 Hz. Training and evaluating SNN object detectors on extended event sequences improves their ability to exploit temporal cues while preserving the energy-efficiency benefits of sparse spiking computation. The results highlight sequence-aware training as a complementary direction to architectural improvements for event-based SNN detection.

**Keywords** Event camera · Spiking neural network · Object detection · Event-based vision

## Introduction

Particularly in highly dynamic scenes, standard frame-based cameras reveal their poor performance in terms of motion blur and nighttime sensing. Their static frame rate further hinders capturing rapid movements of cars and trains, which can cover meters in mere milliseconds. This poses a critical limitation, especially in autonomous driving scenarios where such delays may result in fatal consequences. Therefore, event cameras recently emerged, which follow a retina-inspired sensing principle: each pixel asynchronously reports local logarithmic intensity changes instead of absolute intensity frames [1, 2] in order to achieve a high temporal resolution in the magnitude of microseconds. They further cover a high dynamic range (about 140 dB compared to 60 dB by standard cameras), which allows the detection of movements even in low light settings. Similar to biological retinas, event camera pixels adapt to very dark as well as very bright stimuli, which helps explain their robustness under changing illumination [2]. Combined with their low power consumption, this makes event cameras highly suitable for the usage in automotive edge applications.

However, the inherent sparsity and temporal nature of the event data make it difficult to apply traditional frame-based object detection approaches. A more natural neuromorphic approach to processing events involves the use of Spiking Neural Networks (SNNs) [3], which are motivated by neural computation in the brain: neurons exchange sparse spike events,

✉ Katharina Bendig
katharina.bendig@dfki.de

René Schuster
rene.schuster@dfki.de

Didier Stricker
didier.stricker@dfki.de

[1] RPTU – Technical University Kaiserslautern-Landau, Gottlieb-Daimler-Straße 47, Kaiserslautern 67663, Germany

[2] DFKI – German Research Center for Artificial Intelligence, Trippstadter Str. 122, Kaiserslautern 67663, Germany

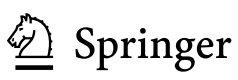

integrate incoming signals in a membrane potential, leak this state over time, and emit an output spike only when a threshold is reached. This makes memory and computation local to the neuron dynamics, following the neuromorphic view that processing and memory can be coupled rather than treated as separate operations [4, 5]. Consequently, SNNs can replace multiplications in a traditional Articial Neural Network (ANN) with additions or simple memory readouts, resulting in a reduced number of machine operations. This makes SNNs very compelling as both energy-efficient networks and stateful models for event-driven perception at a time when energy consumption is a critical issue, particularly in autonomous vehicles.

Real-world scenarios demand continuous data stream processing, where perception benefits from integrating recent evidence instead of repeatedly solving independent snapshot problems. SNNs provide this temporal integration through the membrane potential of each spiking neuron, a brain-inspired internal state for local memory and processing in neuromorphic systems [6]. If no input current arrives, this potential decays over time, which means that SNNs are able to naturally handle cases in which no events are detected. However, current spiking object detectors, like the one by Cordone et al. [7], only consider the time steps in a single time interval and then reset the internal state of the SNNs after one prediction, which is illustrated in Fig. 1. Instead, we propose **Sequence-SOD**, our sequence-aware approach trained on extended sequences encompassing multiple labels across different time points in event data. This method preserves the spiking neurons' inner states within a sequence, keeping the neural memory mechanism active while the event stream evolves, as depicted in Fig. 1b. The internal state is reset only at the boundary between independent sequences. Theoretically, the sequential processing of 25 ms time steps enables bounding-box predictions at a frequency of about 40 Hz.

The main contribution of this paper is a sequence-aware training and evaluation strategy for SNN-based object detection that explicitly treats membrane potentials as the model's temporal memory for continuous perception. In particular:

- We introduce a sequence-aware training approach for object detection using SNNs. In contrast to single-interval SNN detectors, our approach considers whole event sequences that may contain multiple labels at different points in time and maintains the membrane state across the intervals of one sequence, leveraging leaky integration and decay instead of resetting this state after each prediction.
- We analyze the effect of this sequence-aware setting on real-world automotive event data and show that it improves detection performance over single-interval SNN training with the same SSD-based Spiking DenseNet architecture.
- We further evaluate event-data augmentation and compare the estimated inference energy consumption of our SNN architecture with the corresponding ANN.

## Related Work

### Spiking Neural Networks

The work by Maass and Markram [8] has shown that in theory SNNs have exactly the same expressive power as ANNs. In addition, they are not only more energy efficient but also show a greater potential in handling event data compared to ANNs [9].

When modelling SNNs, the most common design is the Leaky Integrate-and-Fire (LIF) neuron model, whose discretized variant can be described in the following way [10]:

$$U[t] = \underbrace{\beta U[t-1]}_{\text{decay}} + \underbrace{WX[t]}_{\text{input}} - \underbrace{S_{out}[t-1]\theta}_{\text{reset}}. \tag{1}$$

The spiking functionality integrates the weighted sum of the incoming spikes $WX[t]$ into the membrane potential $U$ of the spiking neuron. $U$ exponentially decays with a rate $\beta$ to decrease the influence of temporally uncorrelated data.

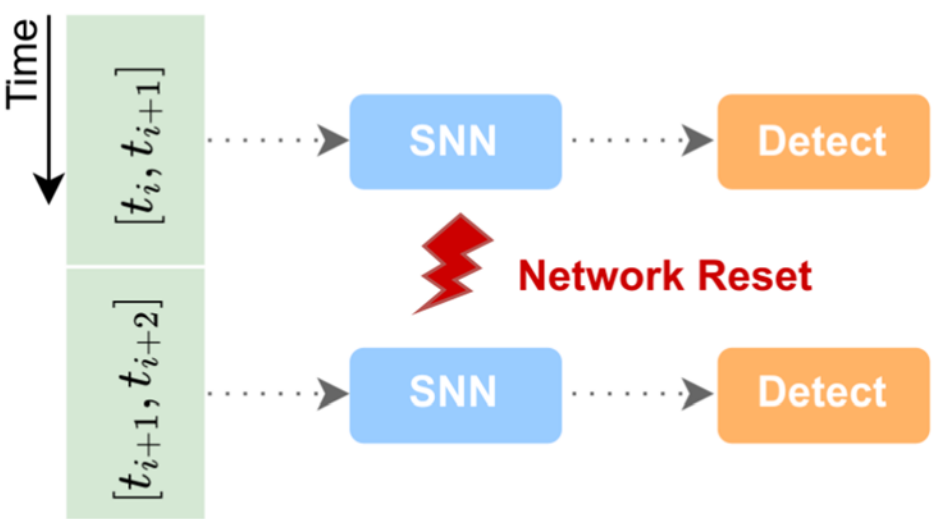

(a) Resetting the network state between samples.

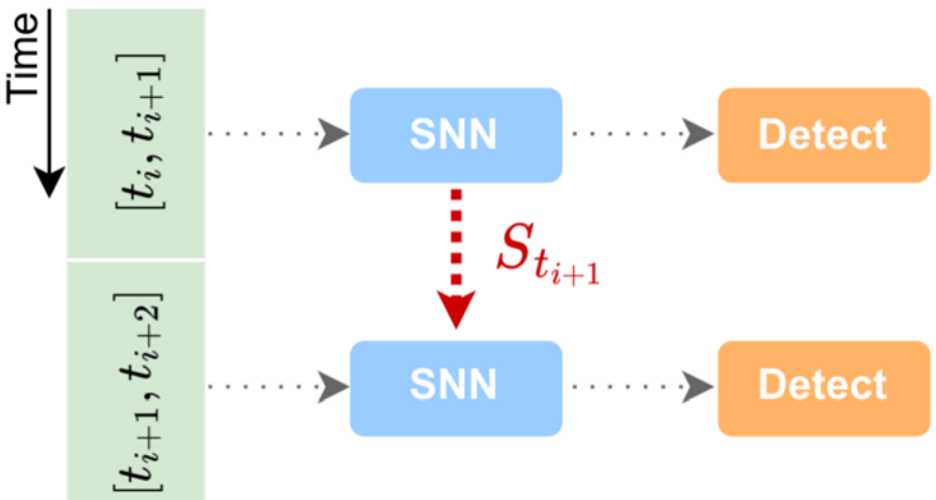

(b) **Sequence-SOD**: Maintaining the network's inner state within a sequence.

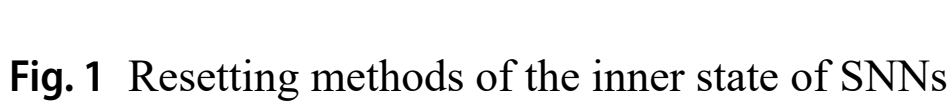

**Fig. 1** Resetting methods of the inner state of SNNs

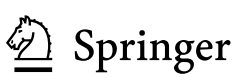

When the membrane potential reaches a certain threshold $\theta$ an output spike is generated and $U$ is reset accordingly.

The disadvantage of this spiking behavior is that it is non-differentiable and therefore the common backpropagation algorithms from ANN training are not directly transferable. So far, three alternative approaches for the challenging training of SNNs have been proposed: Biology-inspired training, conversion from ANNs, and direct training.

Biologically plausible methods, such as Spiketiming-dependent plasticity (STDP) [11], adjust the weights of synapses based on the correlation of the spike behavior of pre- and post-synaptic neurons. However, the locality of updates makes it difficult to train deep networks, rendering biological approaches insufficient for complex tasks such as object detection.

Another option is the conversion of a pre-trained ANN into a SNN by weight-normalization, as there is a correlation between the ReLU activation and the firing rate of a spiking neuron [12]. However, this requires the input to be rate encoded, which means that the number of spikes has to be proportional to the pixel intensity. Consequently, achieving approximately the same accuracy as the pre-trained ANN requires very high latency, defeating the goal of low energy consumption [13]. Furthermore, event data is not suitable as input for ANNs due to its inherent sparsity and asynchronous nature.

The most promising approach is direct training using surrogate gradients [14]. To overcome the non-differentiability of the binary spikes, the gradient is approximated during the backward pass. This also has the advantage of defining a small gradient even when no spike occurs, ensuring the presence of a training signal in such scenarios.

## Event-Data Object Detection

### ANN-Based Object Detection

Event-based object detection has been addressed with a broad range of input representations and network architectures. The sparsity, high temporal resolution, and asynchronous nature of event data make this task challenging for conventional frame-based detectors, but they also motivate architectures that explicitly preserve temporal information.

A common strategy is to convert the asynchronous event stream into dense tensors that can be processed by conventional ANNs, for example through event frames that accumulate events over a fixed time window or event count [15, 16], time-surface representations that encode recent event timestamps at each pixel [17], or learned/custom accumulation layers that transform events into convolution-friendly inputs [18]. Such representations make event data compatible with mature feed-forward detectors, but the resulting models typically process isolated temporal windows. Consequently, temporal information from earlier parts of the stream may be lost unless the detector includes an explicit recurrent, attention-based, or otherwise stateful mechanism.

Recent ANN-based detectors have increasingly focused on temporal memory and efficient sequence modelling for event-camera object detection. The Recurrent Event-camera Detector (RED) [19] combines convolutional layers with ConvLSTM units and demonstrates the benefit of recurrent state modelling for high-resolution event-camera detection. ASTMNet [20] further exploits spatio-temporal recurrence and attention. SODFormer [21] formulates streaming object detection from events and frames as an end-to-end sequence prediction problem with asynchronous attention-based fusion. GET [22] introduces group tokens that decouple temporal-polarity information from spatial information within a Transformer backbone. Zubic et al. [23] propose State Space Models (SSMs) with learnable time-scale parameters for event cameras, enabling adaptation to different temporal scales without retraining. RVT [24] combines convolutional priors, local and global self-attention, and recurrent temporal aggregation to achieve strong accuracy–latency trade-offs. Building on this line of work, SAST [25] applies scene-adaptive sparsification to an RVT-style window-based Transformer while retaining recurrent temporal processing. SMamba [26] combines sparse Mamba-style modelling with ConvLSTM modules for spatio-temporal information propagation, and EvRT-DETR [27] augments RT-DETR [28] with lightweight ConvLSTM-based memory modules to handle event-frame sequences. Together, these works show that retaining, selecting, and propagating temporal information remains central to accurate and efficient event-based object detection.

### SNN-Based Object Detection

SNNs are a natural fit for event-camera data because they process sparse binary spikes and maintain a temporal state through their membrane potentials. They handle small time steps of event data sequentially, allowing a high prediction frequency. Early SNN-based detectors for automotive event data include the DenseNet-based SSD detector of Cordone et al. [7], which demonstrated that directly trained spiking networks can be applied to event-based object detection. More recent work has aimed to improve the accuracy, feature representation, and training stability of SNN detectors. Su et al. [29] propose a directly trained spiking YOLO-style detector, while SFOD [30] introduces multi-scale spiking feature fusion for event-camera object detection. Spiking CenterNet [31] combines a CenterNet-style spiking detector with knowledge distillation, and SpikingViT [32] incorporates spiking computation into a multi-scale Vision Transformer architecture.

Several recent methods further investigate how event representation, temporal processing, and hybrid architectures can improve SNN-based detection. SpikeFPN [33] introduces a spiking feature pyramid network with an adaptive threshold mechanism in selected spiking modules. Hybrid SNN–ANN designs [34] combine spiking layers with conventional ANN components, using attention-based bridge modules to transform sparse spatio-temporal spike features into dense feature maps for the non-spiking detector stages. Hybrid spiking Transformer models such as HsVT [35] similarly combine spiking computation with Transformer-style spatial and temporal feature extraction to capture both local event structure and longer-range temporal dependencies. SpikeYOLO [36] simplifies YOLO-style detection for spiking computation and introduces integer-valued training with spike-driven inference, while Trainable Spiking-YOLO [37] uses trainable spiking detection modules and membrane-potential decoding to improve low-latency object detection. Together, these methods show that stronger SNN object-detection architectures and training strategies can substantially reduce the performance gap to conventional detectors.

Two closely related developments to our work are SpikingViT [32] and STMOD [38], both of which further emphasize the importance of temporal processing for event-based SNN detection. SpikingViT studies the effect of different input lengths for a multi-scale spiking Vision Transformer with a YOLOX-style detection head, while STMOD proposes a spike tensor-based multi-scale detector with a cross-attention spatio-temporal spike generator, an SNN backbone, and an asymmetric decoupled detection head. Since both methods combine spiking modules with non-spiking detector components, we regard them as hybrid SNN-based architectures rather than strictly fully spiking end-to-end detectors. Their temporal ablations and streaming-oriented design choices support our motivation, but their results are coupled with substantial architectural changes; in contrast, our work isolates sequence-aware training and evaluation in an SSD-style Spiking DenseNet by preserving membrane states across consecutive event intervals with multiple possible label timestamps.

Overall, recent SNN-based and hybrid SNN-based detectors mainly focus on architectural design, event encoding, feature fusion, neuron models, detection heads, or training strategies. In contrast, our work studies how an existing fully spiking detector can exploit the temporal continuity of event streams through sequence-aware training and evaluation. This is particularly relevant for real-world event-camera applications, where detectors operate on continuous streams and can use spiking neuron states as temporal memory across labeled and unlabeled intervals.

## Method

Seeking to fully leverage the temporal resolution of event data and the stateful dynamics of spiking neurons [6], our object detection approach processes entire sequences of events containing labels at different time steps, rather than processing individual event samples one at a time. In our approach, events are initially accumulated into short time intervals. Each event interval is then divided into 5 equally-sized time steps which are fed to the network sequentially. The network output belonging to one interval is averaged over these 5 internal time steps for the bounding-box and class prediction. The detection loss is applied to this averaged interval-level prediction when a label is available for the corresponding interval. Thus, the same label is not used as five independent training targets. Notably, the membrane potentials of the spiking neurons are not reset between time intervals within one sequence, which lets the detector retain useful recent evidence and discard outdated information through membrane leakage. The state is reset only between independent sequences.

### Sequential Event Data Processing

The pixels of an event camera register logarithmic intensity changes in an asynchronous manner. An event occurring at the pixel position $(x_i, y_i)$ at time $t_i$ is commonly defined as a tuple $e_i = (x_i, y_i, t_i, p_i)$. Depending on whether the brightness increases or decreases, events have either a positive or negative polarity $p_i \in \{1, -1\}$.

Due to the challenges associated with training deep SNNs on neuromorphic hardware, we simulate the computations in the form of time steps on a Graphics Processing Unit (GPU). Consequently, processing event data asynchronously becomes computationally infeasible, leading us to accumulate events in short time intervals $\Delta t = [t_a, t_b)$ of 125 ms.

Inspired by the event data handling by Gehrig and Scaramuzza [24], we choose a simple pre-processing method creating event histograms $E$ in the shape $(T, 2, H, W)$ with $T$ representing the number of time steps discretized from $\Delta t$ and $(H, W)$ denoting the height and width of the event sensor. We choose to not further divide the time steps into time bins, since Cordone et al. [7] showed that time bins do not improve the performance for $T \geq 5$. Moreover, we accumulate events of different polarities separately in two channels. A stream of events $\mathcal{E}$ is thus processed in the following way:

$$E(\tau, p, x, y) = \sum_{e_i \in \mathcal{E}} \delta(\tau - \tau_i) \cdot \delta(p - p_i) \cdot \delta(x - x_i) \cdot \delta(y - y_i), \tag{2}$$

$$\tau_i = \left\lfloor \frac{t_i - t_a}{t_b - t_a} \cdot T \right\rfloor, \tag{3}$$

with $\delta(\cdot)$ as the Kronecker delta function. Equation 2 assigns each event to exactly one histogram entry according to its discrete time index, polarity, and spatial position. Equation 3 maps the continuous timestamp $t_i$ from the interval $[t_a, t_b)$ to a discrete index $\tau_i \in \{0, \ldots, T-1\}$ by subtracting the interval start time, normalizing by the interval duration, multiplying by $T$, and applying the floor operation.

Additionally, we consider whole sequences of the event stream during training with a sequence length of $s$, which is visualized in Fig. 2. One training sequence includes $s$ time intervals, each containing an event histogram with $T = 5$ time steps over $\Delta t = 125$ ms. Since bounding-box labels in our selected dataset are generated based on the corresponding RGB images, they are intermittently available rather than continuously present, appearing at different points in time with varying frequencies. Consequently, not all time intervals within an event sequence contain bounding boxes. We ensure that at least one label is present in each training sequence. For unlabeled intervals, the network state is propagated forward, but no detection loss is computed. For labeled intervals, the output is accumulated over the $T$ internal time steps and averaged before the classification and localization losses are applied. This creates an interval-level prediction aligned with the available label rather than duplicating the same label across all internal time steps.

The inner state of an SNN, reflected in the membrane potential of its neurons, acts as a compact memory trace of prior event inputs. This stored information, temporally correlated with subsequent inputs, contributes to future predictions. Unlike prior spiking object detection methods that reset the membrane potential after each label occurrence, our SNN maintains its inner state throughout the processing of one sequence and resets it only between independent sequences. Previous methods failed to harness the full potential of temporal information within event data. Processing whole sequences compels the network to prioritize essential temporal cues, which connects the cognitive motivation of stateful perception directly to the detection task. Individual spiking neurons have a finite capacity and as the input sequence lengthens, they must process an increasing volume of incoming data over time. Consequently, the network is forced to learn a more meaningful inner state representation of the current scenario, discarding redundant information to process new stimuli. Multiple labels at different points in time prompt the network to generate accurate predictions not only at the sequence's end but also at intermediate labeled intervals. This compels the network to consistently maintain an expressive inner state representation, facilitating ongoing predictions in a continuous stream of information.

## Network Architecture

For consistent comparability, our approach follows the methodology of Cordone et al. [7] by adopting the SSD object detection framework [39] with a Spiking DenseNet (depth of 121 and a growth rate of 24) as its backbone. The dense connectivity pattern of DenseNet [40] enhances the propagation of gradients, which is especially important for SNNs since the gradient can only be approximated. Following the example of Cordone et al. [7], we replace the ReLU activations with Parametric Leaky Integrate-and-Fire (PLIF) [41] neurons, which in contrast to conventional LIF neurons include the decay rate $\beta$ as a trainable parameter. This choice keeps the architecture comparable to Cordone et al. [7] while allowing the membrane time constant to adapt to the temporal statistics of the event stream, so that different neurons can learn how quickly they should retain or forget recent evidence.

The DenseNet architecture includes batch normalization, which seems to play a crucial role in the training of our SNN. However, after training, it is possible to construct a mathematically equivalent network that absorbs the batch normalization parameters into the convolutional layers [42]. Therefore, our network can be seen as a pure SNN during inference, leveraging all the advantages in terms of energy efficiency of the spiking computations.

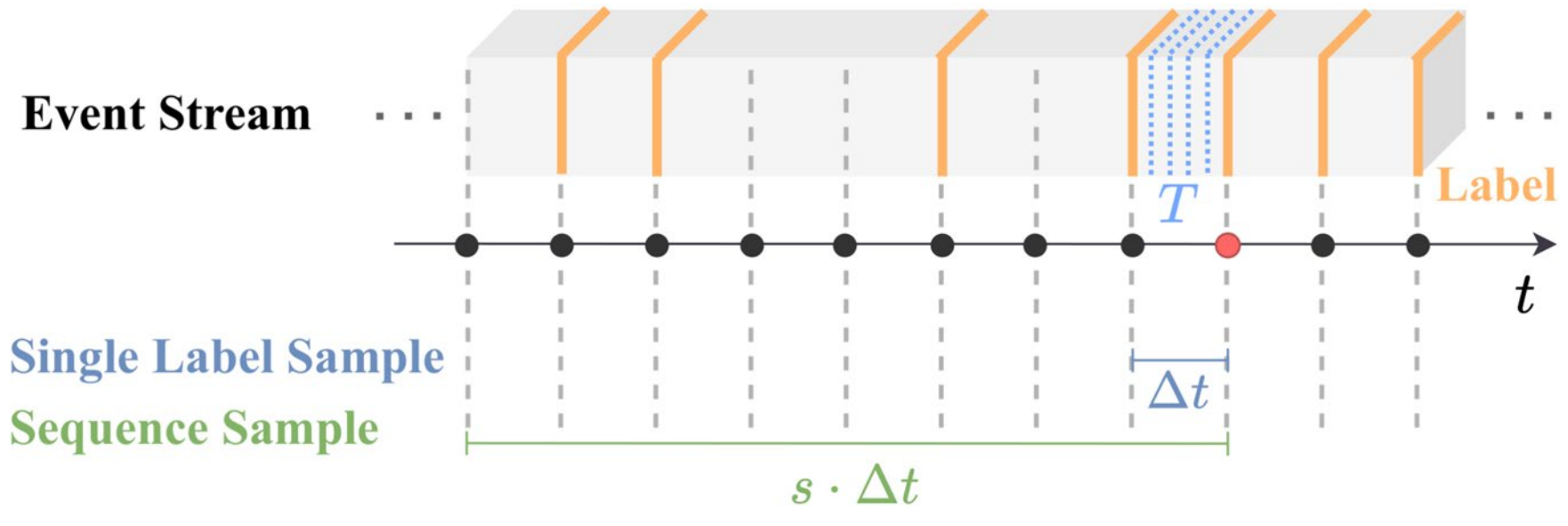

**Fig. 2** Visualization of the sampling strategies. Black markers denote consecutive event intervals. The blue bracket illustrates a single-label sample, which consists of one event interval $\Delta t$, whereas the green bracket illustrates a sequence sample of length $s$, which contains $s$ consecutive event intervals with multiple labels. Orange markers indicate time points at which bounding-box labels are available. The red marker highlights one exemplary labeled interval. The figure is schematic; unless stated otherwise, the experiments use $s_{train} = 5$, $\Delta t = 125$ ms, and $T = 5$

In contrast to previous recurrent object detectors [19, 20, 24], which process multiple time steps jointly, our spiking approach processes an input tensor of shape (2, $H$, $W$) at each time step. The final prediction of the bounding boxes and classes for one interval is calculated by accumulating the network output over $T$ time steps and dividing it by $T$. During training, gradients are back-propagated through the unrolled sequence of SNN states. The non-differentiable spike function is handled using surrogate gradients, following direct SNN training. Thus, the loss at labeled intervals updates both spatial weights and temporal membrane-state dynamics across preceding intervals. Beyond the initial five time steps (equivalent to 125 ms), the output accumulation can be performed with a sliding window approach for the rest of the sequence. Consequently, the network is theoretically able to output bounding-box predictions after every 25 ms time step (= 40 Hz), effectively leveraging the high temporal resolution of the event camera. However, the prediction frequency relies on the input data and may improve with smaller time steps, albeit potentially impacting performance negatively.

## Experiments and Results

### Setup

**Dataset** As we focus on object detection on event data in an automotive setting, we test our method on the Gen1 Automotive Detection Dataset [43]. The dataset contains 39 h of event camera recordings from diverse driving scenarios with a resolution of 304 px × 240 px. Its 255 k labels consist of bounding boxes for pedestrians and cars provided at frequencies between 1 and 4 Hz. In the fashion of previous work [19, 20, 24], we do not consider small bounding boxes with any side length $\leq$ 10 px or a diagonal $\leq$ 30 px. The smallest temporal gap between two bounding-box annotations in the dataset is 250 ms. We therefore construct sequence samples every 125 ms, i.e. at half of this smallest label interval, to align the event-interval sampling grid with possible label timestamps while still propagating the SNN state through intermediate unlabeled intervals. For example, two consecutive labels separated by 250 ms can be connected by two 125 ms event intervals. The first interval updates the membrane state, and the interval associated with an available label receives the detection loss.

Since it is a widely used metric for object detection, we report the mean Average Precision (mAP) over 10 IoU thresholds with the starting value 0.5, an increment of 0.05, and an ending value of 0.95. $AP_{50}$ and $AP_{75}$ denote Average Precision at IoU thresholds of 0.50 and 0.75, respectively. $AP_{car}$ and $AP_{ped}$ report class-specific Average Precision for cars and pedestrians.

**Implementation** We adopt standard training strategies based on the implementation of Cordone et al. [7], utilizing an AdamW optimizer with a weight decay of $1 \times 10^{-4}$ and implementing a cosine decay for the learning rate. Additionally, we train with a batch size of 14 and an initial learning rate of $4 \times 10^{-5}$. The batch size is limited by the memory requirements of sequence-based SNN training through time. For the event data input, we consider sequences of length $s_{train} = 5$, where each time interval consists of a discretized 125 ms time window which is further divided into $T = 5$ time steps. Unless stated otherwise, $T$ is fixed to 5 in all experiments. This results in internal time steps of 25 ms and therefore a theoretical prediction frequency of 40 Hz. Just like Cordone et al. [7], we utilize the Focal Loss [44] for the classification task, since it efficiently addresses the class imbalance issue, as well as a smooth L1 loss for the box regression. We evaluate the trained models after 110 epochs and report the best-performing checkpoints according to validation performance. Inspired by the work of Gehrig and Scaramuzza [24], we study the influence of data augmentation, including horizontal flipping and zooming in/out during training, in Section “Data Augmentation”.

The SSD-style post-processing follows the baseline detector and includes Non-Maximum Suppression (NMS). For NMS we use the same parameters as in Cordone et al. [7] to ensure comparability.

### Ablation Study

In this section, we explore the key factors that considerably influence our approach’s performance. We investigate the effect of sequence length during training and assess the influence of event data augmentation on training our SNN. The experiments are conducted on the Gen1 Automotive Detection Dataset [43].

#### Sequence Length

Table 1 shows the results obtained by training our network on a single time interval $s_{train} = 1$ and a longer sequence length $s_{train} = 5$. We test the resulting networks on different sequence lengths $s_{test}$, including *full* sequences of the test split of the Gen1 Automotive Detection Dataset. A sequence is considered *full* if the maximum temporal gap between labels is 200 time intervals. Since one interval corresponds to 125 ms, this limit corresponds to 25 s. Sequences with larger temporal gaps are split in order to avoid excessive computation on long stretches without any label. We focus on $s_{train} = 1$ and $s_{train} = 5$ to compare the prior single-interval setting with our sequence-aware setting while keeping all other architectural choices fixed. The data suggests a performance improvement corresponding to the length of the training sequence. This improvement is observed across both object classes and different IoU thresholds. Training on longer sequences compels the network to prioritize the storage of important temporal information in its inner network state. The capacity of a single spiking neuron is limited, and during the processing of larger amounts of data, the network has to restrict the information flow in order to achieve the desired firing of the output neurons. A longer sequence can also contain multiple labels at different points in time, pushing the network to output correct predictions throughout the sequence. Therefore, even when tested on short sequences the network trained with $s_{train} = 5$ performs better. The data further shows that the performance of the $s_{train} = 1$ training decreases slightly when tested on the *full* sequence, indicating that the network is not able to fully leverage past temporal information because its membrane potentials are reset during training. Especially, its prediction for pedestrians dropped, posing a harder challenge due to more variety in movement. Figure 3 visualizes the results for the networks trained on sequence lengths 1 and 5 tested on *full* sequences. It can be seen that the network trained on longer sequences is more capable of detecting multiple entities even with a small number of events in one time interval, as in Fig. 3a. Furthermore, it better differentiates between multiple object instances due to past temporal cues, as demonstrated in Fig. 3d.

**Table 1** Evaluation of different training sequence lengths $s_{train}$ on various test sequence lengths $s_{test}$ of the Gen1 Automotive Detection Dataset.

| $s_{train}$ | $s_{test}$ | $\Delta t$ in ms | mAP | $AP_{50}$ | $AP_{75}$ | $AP_{car}$ | $AP_{ped}$ |
|---|---|---|---|---|---|---|---|
| 1 | 1 | 125 | 23.38 | 47.23 | 20.97 | 35.39 | 11.36 |
| 1 | 5 | 125 | 23.63 | 47.74 | 21.02 | 35.42 | 11.83 |
| 1 | full | 125 | 23.04 | 45.89 | 20.68 | 35.17 | 10.92 |
| 5 | 1 | 125 | 25.26 | 50.98 | 21.88 | 36.09 | 14.43 |
| 5 | 5 | 125 | 25.31 | 50.99 | 21.89 | 36.05 | 14.57 |
| 5 | full | 125 | 25.30 | 51.49 | 21.91 | 35.87 | 14.72 |

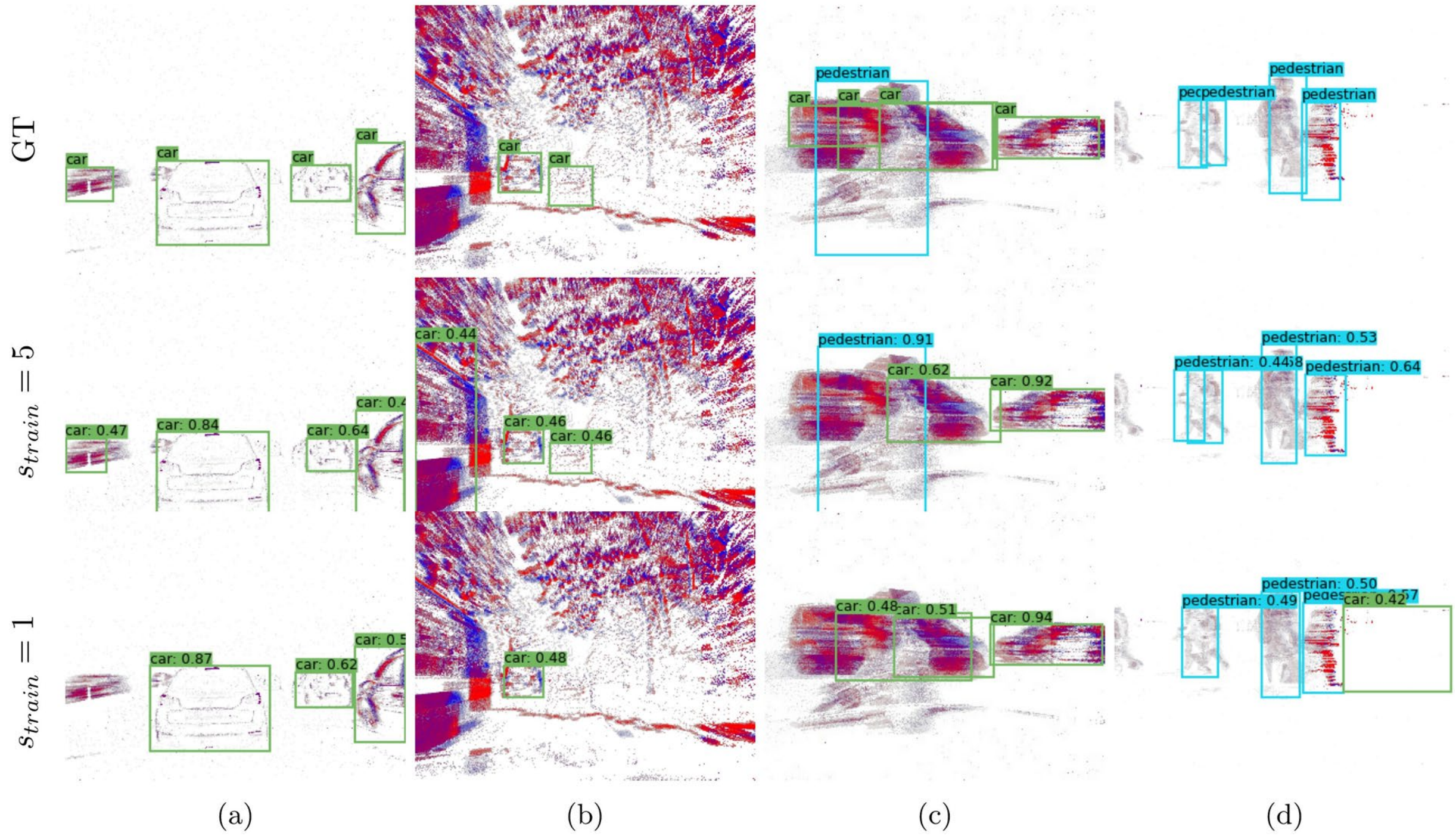

**Fig. 3** Visualization of the predictions on the Gen1 Automotive Detection Dataset [43] with different training sequence lengths along with the ground truth bounding boxes, showing the network behavior in different scenarios (varying density of events and varying number of objects)

### Data Augmentation

Following the example of Gehrig and Scaramuzza [24], we examine during training the effect of the following randomized event data augmentations on the performance of our SNN: horizontal flipping, zooming in, and zooming out. We randomly choose whether to use no augmentation, or flipping and/or zooming, and then apply this choice to a whole sequence. For any resizing operation we use exact nearest neighbor interpolation in order to keep the data in its binary format. To zoom in, we crop the sample randomly along its spatial dimensions and then resize it to the original resolution. For zooming out, we down-sample the whole sequence spatially and place it at a random location within a zero tensor of the original size. Our parameter choice with respect to the randomness follows the one of Gehrig and Scaramuzza [24]. Table 2 shows the results of the different augmentation methods applied to the training based on sequence length 5. We can see that augmentation plays an important role for the training of SNNs. Zooming is most effective, especially for pedestrians, which usually appear smaller than cars in real world data, possibly due to an increased focus on particular features of otherwise small objects. Zoom-in alone achieves the highest $AP_{50}$ and $AP_{ped}$, while the combination of all three augmentations yields the highest overall mAP and $AP_{75}$. This indicates that zoom-in is particularly beneficial for small-object recognition, whereas the combined augmentation improves robustness at stricter localization thresholds. All reported test results are obtained without applying these augmentations during testing.

### Benchmark Comparison

In this section, we contrast our sequence-aware SNN approach **Sequence-SOD** with previous event-based object detection methods, evaluating their performance on the Gen1 Automotive Detection Dataset. A summary of the results is presented in Table 3. The table includes a broad comparison of ANN-based, hybrid, and SNN-based detectors; however, the most important comparison for our study is Cordone et al. [7], which uses the same SSD-style Spiking DenseNet architecture but is trained on isolated event intervals. Testing the benefits of our sequence-aware training and evaluation strategy on other detector architectures remains future work.

The results show that methods with explicit temporal modelling, such as recurrent, Transformer-based, and hybrid architectures, generally achieve substantially higher Gen1 accuracy than early feed-forward event-frame detectors. Hybrid SNN–ANN approaches and hybrid spiking Transformer models also report strong results, suggesting that combining spiking computation with non-spiking modules can improve detection accuracy. Recent SNN-based detectors further narrow the gap by improving event encoding, feature fusion, neuron models, and detection heads. These trends confirm that architecture remains an important factor, but they are largely orthogonal to the question studied here.

In our work, we demonstrate the importance of a temporal inner state for SNNs. This can be seen by comparing our results to ODSNN [7], which resets the spiking state after isolated event intervals. The numbers indicate that training on longer sequences improves the performance by enforcing a meaningful inner state representation, which is able to discard dispensable information and prioritize significant temporal cues. Without augmentation, sequence-aware training improves mAP from 23.38 to 25.30, showing that the sequence formulation itself contributes to the gain. With augmentation, the result increases further to 26.88. We therefore separate the effect of sequence-aware training from the effect of data augmentation.

Compared with more recent SNN-based detectors such as EMS-YOLO [29], SFOD [30], and SpikeFPN [33], Sequence-SOD is not intended to compete primarily through architectural novelty. These methods improve the detector backbone, feature fusion, or detection head, whereas our architecture intentionally follows Cordone et al. [7] to isolate the effect of preserving temporal state during sequence-aware training and evaluation.

### Energy Consumption

Apart from their naturally suitable ability to process event data, a very significant advantage of SNNs is in terms of energy efficiency. In comparison to traditional ANNs, in which all neurons are activated for every iteration, SNNs have sparse activations. Furthermore, Multiply and Accumulate (MAC) operations for the synaptic weights are replaced by Accumulate (AC) operations in SNNs, since the spikes are represented only by zeros and ones.

**Table 2** Application of geometrical augmentation methods for the training on sequence length 5 and tested on *full* sequences

| h-flip | zoom-in | zoom-out | mAP | $AP_{50}$ | $AP_{75}$ | $AP_{car}$ | $AP_{ped}$ |
|---|---|---|---|---|---|---|---|
| | | | 25.30 | 51.49 | 21.91 | 35.87 | 14.72 |
| ✓ | | | 25.40 | 51.03 | 22.11 | 36.19 | 14.61 |
| | ✓ | | 26.85 | 53.41 | 23.63 | 37.80 | 15.90 |
| | | ✓ | 26.29 | 52.75 | 23.18 | 37.66 | 14.93 |
| ✓ | ✓ | ✓ | 26.88 | 52.79 | 24.65 | 37.90 | 15.86 |

**Table 3** Comparison to State-of-the-Art on the test split of the Gen1 Automotive Detection Dataset. Information that is approximated, does not apply, or is not available is marked with a star *. The results for the ◇ marked papers were taken from Gehrig and Scaramuzza [24]

| Method | Arch. | Det. Head | Seq | Aug | $\Delta t$ | mAP | Params (M) |
|---|---|---|---|---|---|---|---|
| Inception + SSD [45]◇ | CNN | SSD | ✗ | – | – | 30.1 | > 60* |
| RRC-Events [46]◇ | CNN | YOLOv3 [47] | ✗ | – | – | 30.7 | > 100* |
| YOLOv3 Events [48]◇ | CNN | YOLOv3 [47] | ✗ | – | – | 31.2 | > 60* |
| RED [19] | RNN | SSD | ✓ | ✗ | 50 | 40.00 | 24.1 |
| ASTMNet [20] | RNN | SSD | ✓ | ✗ | * | 46.7 | > 100* |
| RVT-B [24] | RNN/Trans. | YOLOX [49] | ✓ | ✓ | 50 | 47.2 | 18.5 |
| RVT-S [24] | RNN/Trans. | YOLOX [49] | ✓ | ✓ | 50 | 46.5 | 9.9 |
| Hybrid SNN–ANN [34] | Hybrid/RNN | YOLOX [49] | ✓ | ✓ | 50 | 43.0 | 7.7 |
| HsVT-B [35] | Hybrid/Trans. | Head* | ✓ | * | 50 | 47.8 | 17.2 |
| EMS-YOLO [29] | SNN | YOLOv3 [47] | ✗ | ✗ | * | 26.7 | 6.2 |
| Spiking CenterNet [31] | SNN | CenterNet | ✗ | * | 100 | 22.9 | 13.0 |
| SpikeFPN [33] | SNN | FPN | ✗ | * | 60 | 22.3 | 22.0 |
| STMOD [38] | Hybrid/SNN | ADH | ✓ | * | 5 | 44.3 | 6.9 |
| SFOD [30] | SNN | SSD | ✗ | ✓ | 100 | 32.1 | 11.9 |
| SpikingViT [32] | Hybrid/SNN/Trans. | YOLOX [49] | ✓ | ✓ | 50 | 39.4 | 21.5 |
| SpikeYOLO [36] | SNN | YOLO head | ✗ | * | 2500 | 40.4 | 23.1 |
| ODSNN [7] (Baseline) | SNN | SSD | ✗ | ✗ | 100 | 18.9 | 8.2 |
| S-SOD (ours) | SNN | SSD | ✗ | ✗ | 125 | 23.38 | 8.2 |
| S-SOD (ours) | SNN | SSD | ✓ | ✗ | 125 | 25.30 | 8.2 |
| S-SOD (ours) | SNN | SSD | ✓ | ✓ | 125 | 26.88 | 8.2 |

Following the convention of previous work [50], we determine the energy consumption in terms of the total number of Floating point operations (FLOPs) on standard CMOS technology. For a standard ANN, the FLOPs for multiplication and accumulation in a convolutional layer are both calculated by:

$$FLOPs^{Mult}_{conv,l} = FLOPs^{Ac}_{conv,l} = k^2 \times H_{out} \times W_{out} \times C_{in} \times C_{out}, \quad (4)$$

with $k$ as the kernel size, $(H_{out}, W_{out})$ as the output feature map height and width as well as $C_{in}, C_{out}$ as the number of input and output channels respectively.

Since spiking neurons only consume energy when a spike occurs, their FLOPs of a layer $l$ reduce based on the spike rate $R_l$. An average spike rate can be computed as the total number of spikes in a layer for all time steps divided by the number of neurons in $l$:

$$R_l = \frac{\sum_{t\in\{1,\dots,T\}, n\in N_l} S^n_t}{|N_l|}, \quad (5)$$

with $N_l$ as the set of neurons in layer $l$ and $S^n_t$ as the binary spike if neuron $n$ fired at time step $t$. The resulting spike rates per layer for our **Sequence-SOD** with $s_{train} = 5$ and $s_{test} = 1$ on the Gen1 Automotive Detection Dataset are shown in Fig. 4. This figure is important for the energy analysis because the effective number of SNN operations is directly scaled by the measured spike rate. Averages over entire parts of the network are presented in Table 4.

Additionally, our network contains maxpool layers, which do not perform traditional floating-point operations. We estimate their FLOPs for ANNs and SNNs as an accumulation operation in the following way:

$$FLOPs^{Ac}_{mp,l} = k^2 \times H_{out} \times W_{out} \times C_{in}. \quad (6)$$

We can now obtain the overall inference energy of an ANN ($E^{ANN}$) and of the equivalent SNN ($E^{SNN}$) across all layers $l$ of a network $N$:

$$E^{ANN} = \sum_{l\in N} FLOPs^{Mult}_{conv,l} \times E_{Mult} + \left(FLOPs^{Ac}_{conv,l} + FLOPs^{Ac}_{mp,l}\right) \times E_{Ac} \quad (7)$$

$$E^{SNN} = \sum_{l\in N} \left(R_l \times FLOPs^{Ac}_{conv,l} + FLOPs^{Ac}_{mp,l}\right) \times E_{Ac}, \quad (8)$$

where $E_{Mult} = 3.7$ pJ and $E_{Ac} = 0.9$ pJ are the energy consumed by a 45 nm CMOS process for a 32-bit floating-point multiplication and accumulation operations respectively [51]. The results of our analysis can be found in Table 4, showing a large estimated energy advantage of the SNN compared to the equivalent ANN. The ANN and SNN have the same underlying convolutional topology, but the ANN requires both multiplication and accumulation operations for each convolutional operation. In the SNN, binary spikes remove the multiplication by the activation value, and the effective accumulation operations are scaled by the spike rate $R_l$. Thus, the table reports the same architectural accumulation count for the equivalent network but a reduced effective SNN energy due to sparse spiking activity. It is important to note however, that the energy consumption of SNNs depends on the specific dataset and the resulting spike rates in contrast to ANNs, which always execute the same number of operations. Moreover, SNNs can only achieve this level of energy efficiency, when deployed on neuromorphic hardware.

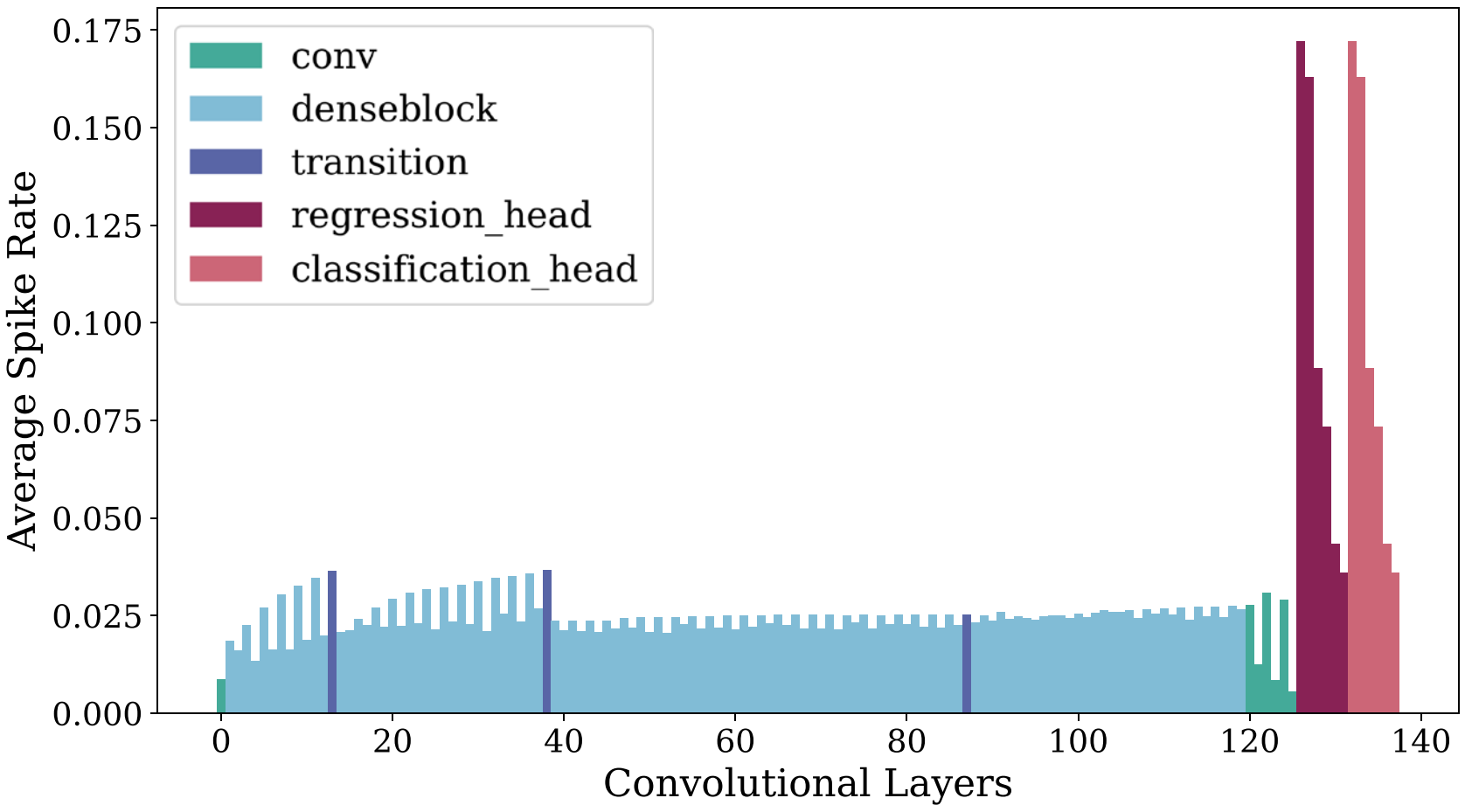

**Fig. 4** Average Spike Rates over all time steps for each convolutional layer

## Discussion and Limitations

The training of SNNs on GPUs poses a great challenge, since GPUs are based on synchronous and parallel computing while the operations in SNNs are event-based. Therefore, the training can only be simulated using time steps and enforces a certain degree of accumulation of events into a denser representation. This makes it especially difficult to train on long sequences of input data, because the gradient has to be computed over all time steps of a whole sequence, leading to a trade-off between training time and performance. Using longer sequences that contain multiple labels may result in a stronger training signal, however this signal will be distributed during backpropagation along the space and time dimensions of the network. Furthermore, SNNs still suffer from a lower accuracy compared to traditional ANNs and Recurrent Neural Networks (RNNs). Our method does not close this gap completely. Instead, it addresses one limitation of previous SNN object detectors, namely the reset of the network state after isolated event samples. Better training methods, neuron models, and detection architectures for SNNs remain important research topics. Still, SNNs can achieve a great reduction in energy requirements compared to equivalent ANN architectures.

This study is also limited to the Gen1 Automotive Detection Dataset and to one SSD-based SNN architecture. We selected this setting to ensure direct comparability with previous SNN object detection work. Validation on additional event-based detection datasets and on further SNN detection architectures would strengthen the generality of the conclusions. In addition, we fixed $T = 5$ and $\Delta t = 125$ ms in all reported experiments. These choices provide 25 ms internal time steps and align well with the Gen1 annotation timing, but a systematic ablation over $T$, $\Delta t$, and intermediate sequence lengths remains future work.

## Conclusion

We introduce a sequence-aware training approach for event-based object detection with SNNs. The approach combines retina-inspired event sensing with brain-inspired spiking dynamics by keeping the leaky membrane potentials of the detector active across consecutive event intervals. Our approach uses longer sequences, which include labels at multiple points in time, during training in order to better mimic continuous real-world usage and the stateful nature of event-driven perception. This allows the network to prioritize the storage of important temporal cues within its internal memory that is represented by the membrane potential of the neurons. This leads to increased detection performance compared with single-interval SNN training under the same architecture. Our network is further able to process not only short, single-label samples, but also longer sequences of continuous event streams with a theoretical prediction frequency of 40 Hz. Future research avenues may explore truncated backpropagation techniques, additional event-based datasets, intermediate sequence lengths, and the application of sequence-aware training to more recent SNN detection architectures in order to determine the full potential of long event sequences while avoiding the computational overhead of the full gradient. Such extensions should preserve the core motivation of sparse, stateful computation while improving scalability and accuracy.

**Table 4** Energy consumption of an ANN and a SNN

| Method | Network | $FLOPs^{Mult}$ | $FLOPs^{Ac}$ | $R_l$ | $E$ (mJ) | $\frac{E^{ANN}}{E^{Method}}$ |
|---|---|---|---|---|---|---|
| ANN | Backbone | 2295.49 M | 2298.21 M | – | 10.56 | 1× |
| | Heads | 351.32 M | 351.32 M | – | 1.61 | 1× |
| | Overall | 2646.81 M | 2649.53 M | – | 12.17 | 1× |
| SNN | Backbone | – | 2298.21 M | 0.0235 | 0.05 | 199× |
| | Heads | – | 351.32 M | 0.0960 | 0.05 | 31× |
| | Overall | – | 2649.53 M | 0.0296 | 0.10 | **116×** |

**Author Contributions** Katharina Bendig implemented and executed the main part of the experiments, contributing significantly to the experimental design and analysis. Katharina Bendig also took the lead in drafting the initial manuscript. René Schuster and Didier Stricker provided critical supervision and guidance throughout the development and refinement of the research idea. Their expertise and insights greatly influenced the conceptualisation of the experiments and the interpretation of results. All authors actively engaged in thorough manuscript review and contributed substantially to its improvement. Their collective feedback, revisions, and valuable input significantly enhanced the overall quality of the manuscript.

**Funding** Open Access funding enabled and organized by Projekt DEAL. This work was funded by the Carl Zeiss Stiftung, Germany under the Sustainable Embedded AI project (P2021-02-009).

**Data Availability** All our experiments as well as the images in Fig. 3 are based on the Gen1 Automotive Detection Dataset, which is available for download from Prophesee at https://www.prophesee.ai/2020/01/24/prophesee-gen1-automotive-detection-dataset/.

## Declarations

**Conflicts of Interest Statement** All authors certify that they have no affiliations with or involvement in any organization or entity with any financial interest or non-financial interest in the subject matter or materials discussed in this manuscript.